# Edge YOLO: Real-Time Intelligent Object Detection System Based on Edge-Cloud Cooperation in Autonomous Vehicles

Siyuan Liang, Hao Wu, *Student Member, IEEE*, Li Zhen, *Member, IEEE*, Qiaozhi Hua, Sahil Garg, *Member, IEEE*, Georges Kaddoum, *Member, IEEE*, Mohammad Mehedi Hassan, *Senior Member, IEEE*, and Keping Yu, *Member, IEEE*

*Abstract*—Driven by the ever-increasing requirements of autonomous vehicles, such as traffic monitoring and driving assistant, deep learning-based object detection (DL-OD) has been increasingly attractive in intelligent transportation systems. However, it is difficult for the existing DL-OD schemes to realize the responsible, cost-saving, and energy-efficient autonomous vehicle systems due to low their inherent defects of low timeliness and high energy consumption. In this paper, we propose an object detection (OD) system based on edge-cloud cooperation and reconstructive convolutional neural networks, which is called Edge YOLO. This system can effectively avoid the excessive dependence on computing power and uneven distribution of cloud computing resources. Specifically, it is a lightweight OD framework realized by combining pruning feature extraction network and compression feature fusion network to enhance the efficiency of multi-scale prediction to the largest extent. In addition, we developed an autonomous driving platform equipped with NVIDIA Jetson for system-level verification. We experimentally demonstrate the reliability and efficiency of Edge YOLO on COCO2017 and KITTI data sets, respectively. According to COCO2017 standard datasets with a speed of 26.6 frames per second (FPS), the results show that the number of parameters in the entire network is only 25.67 MB, while the accuracy (mAP) is up to 47.3%.

Manuscript received February 25, 2021; revised October 5, 2021, November 10, 2021, and January 24, 2022; accepted February 22, 2022. This work was supported by the King Saud University, Riyadh, Saudi Arabia, through the Researchers Supporting Project under Grant RSP 2022/18. The Associate Editor for this article was H. Lu. *(Siyuan Liang and Hao Wu are co-first authors.) (Corresponding authors: Qiaozhi Hua; Sahil Garg.)*

Siyuan Liang, Hao Wu, and Li Zhen are with the Shaanxi Key Laboratory of Information Communication Network and Security, Xi'an University of Posts and Telecommunications, Xi'an 710122, China (e-mail: liangsiyuan@xupt.edu.cn; haowu@stu.xupt.edu.cn; lzhen@xupt.edu.cn).
Qiaozhi Hua is with the Computer School, Hubei University of Arts and Science, Xiangyang 441000, China (e-mail: 11722@hbuas.edu.cn).
Sahil Garg and Georges Kaddoum are with the Electrical Engineering Department, École de Technologie Supérieure, Montréal, QC H3C 1K3, Canada (e-mail: sahil.garg@ieee.org; georges.kaddoum@etsmtl.ca).
Mohammad Mehedi Hassan is with the Department of Information Systems, College of Computer and Information Sciences, King Saud University, Riyadh 11543, Saudi Arabia (e-mail: mmhassan@ksu.edu.sa).
Keping Yu is with the Global Information and Telecommunication Institute, Waseda University, Tokyo 169-8050, Japan (e-mail: keping.yu@aoni.waseda.jp).
In this article, our newly trained model and other comparison models are available on https://github.com/Adam123wu/Edge-YOLO.git.
Digital Object Identifier 10.1109/TITS.2022.3158253

*Index Terms*—Intelligent transportation system, deep learning, object detection, edge-cloud cooperation, convolutional neural networks.

## I. Introduction

**W**ITH ultra-high bandwidth and controllable delay, the emerging 5G technology accelerates the development of the Internet of Things (IoT) [1], [2]. By virtue of 5G future networks, more smart devices at the edge can access to the cloud platform rapidly and reliably [3]. Intelligent Transportation System (ITS) realizes traffic monitoring and drive assistant relying on smart devices that can be deployed in massive machine-type communication (mMTC) scenario of 5G [4]. However, massive amounts of data will be generated in this scenario, especially image and video, which are transmitted to the edge layer of the network [5], [6]. Such big data streams will result in network congestion, delay and other risks which bring more challenges for users in quality of service and quality of experience [7]. In this regard, edge computing is deemed as potential solution, which can compensate for the uneven distribution of computing power of each edge node generated by the cloud due to releasing the restriction of transmission bandwidth [8]. In ITS, the edge layer processing achieves offline availability and makes the devices tend to be miniaturized, which is also privacy protection for users [9], [10]. Vision-based traffic safety monitoring, and driving assistance systems of ITS will be important application scenarios of edge computing based on 5G, which attach great importance to the processing efficiency and consumption [11]. However, with the continuous evolution of artificial intelligence (AI), the complexity of vision processing algorithms and uninterruptible demand for computing resources has increased rapidly, which is an enormous challenge for the existing ITS that rely on object detection [12].

Object detection is one of the four major tasks in the field of computer vision. Currently, the mainstream algorithms in the field of object detection can be divided into two types (i.e, two-stage type and one-stage type). Among those algorithms, the most representative one is the You Only Look Once (YOLO) [13] series of object detection networks. The latest YOLO network neutralizes the high detection accuracy of the two-stage series network and the fast speed of the one-stage network. For devices with high-performance Graphics Processing Units (GPU) such as traditional servers or PCs, the YOLO algorithm can achieve high real-time characteristics and an ideal balance between performance speed and detection accuracy during its continuous evolution. However, for the mobile edge computing-based ITS, intelligent devices often have low power consumption, weak GPU performance, and small Random Access Memory (RAM) capabilities. The existing object detection algorithms cannot meet the corresponding performance requirements. Therefore, it is imperative to carry out lightweight processing for the original network structure of YOLO. Furthermore, in the case that the calculations are performed in intelligent transportation equipment at the edge of the network, the improved YOLO network still needs to maintain advanced real-time performance while ensuring the detection accuracy [14].

In recent years, the embedded platforms for image processing in ITS has developed rapidly, including traditional Field-Programmable Gate Arrays (FPGA), Advanced Reduced Instruction Set Computing Machine, System on Chip (SoC), and the new NVIDIA Jetson [15] series of AI. Unlike traditional PCs with high-performance graphics cards, they can always practically implement AI model calculations at the edge layer, and has the advantages of less power consumption, small space, less heat dissipation. More importantly, they support a variety of types of deep learning frameworks, which enables model reasoning and large-scale deployment in edge computing scenarios. It is worth mentioning that when carrying on mobile platforms such as data-driven based reliable vehicles, robots, and drones, the flexibility will be promoted more obviously. This is undoubtedly an important foothold for AI and edge computing from theory to practice, from independence to interconnection, which can also facilitate the further upgrade of edge computing to edge intelligence (AIoT) in ITS [16].

Given the high computational complexity of existing target detection algorithms, the cloud computing model is widely used in ITS to process various types of traffic flow data. However, it may be inapposite in practical application scenarios due to the following shortcomings: 1) It is difficult to ensure real-time performance, especially for traffic accident detection and assisted driving systems. 2) The resource occupancy rate is high, with the insufficient resources of communication. 3) Excessive energy consumption will be generated, especially for image data transmission. Federated Learning (FL) [17] based on edge-cloud cooperation (E-CC) may be a new solution to solve the above problems, in which each terminal uses locally collected data to train the neural network at the edge, and only weights are uploaded back to the server. Since this strategy aims to solve isolated data islands problems and protect the privacy and data security, the edge layer needs to complete data labeling and dataset training locally. It is unfinished for edge computing devices with weak computing power and high real-time traffic scenarios such as Internet of Vehicles (IoV) [18] and traffic flow detection. Therefore, a more reasonable E-CC solution is needed to hand over the model training to the cloud and conduct model inference in the edge end.

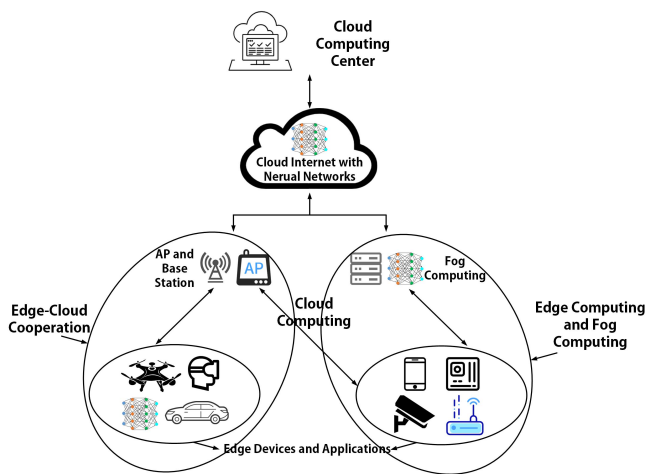

Fig. 1. Three different structures for cloud and edge computing layers and related applications based on neural networks: E-CC, Cloud Computing, edge Computing and Fog Computing.

In this paper, we propose an AIoT system named Edge YOLO, which is more adaptive to existing ITS computing equipment. For the edge layer, we redesign the network structure based on YOLOV4 and propose a more suitable algorithm, Edge YOLO after lightweight processing, which can be preset on the edge computing device. To deal with various scenarios of real-time object detection in ITS, we deploy the neural network directly on the edge node and place the image data collection and storage on the local device, as shown in Fig. 1 When the device is idle, the data will be uploaded, and the cloud is responsible for filtering and labeling the new data collected at the edge. Then, the cloud retrains the data to obtain new weights and updated model parameters, and finally, downstream them to the edge device. Fig. 1 also shows a classic fog computing and cloud computing solution, where fog computing such as small servers with storage is closer to the edge layer, while cloud computing directly accesses edge devices into the cloud computing resource pool. Meanwhile, the edge layer does not perform computing tasks. This solution is suitable for small size devices that need local networking, or many terminals. Therefore, our E-CC-based solution can meet the requirements of mobility and real-time scenarios under the premise of ensuring reliability, which can be used for advanced assisted driving and traffic monitoring in ITS when network resources are not sufficient. To sum up, the majorcontributions of this paper are as follows:

1) *More flexible object detection network*: Based on the latest object detection network YOLOV4 [19] in the YOLO series, the proposed Edge YOLO is more suitable for edge computing scenarios based on 5G for traffic safety monitoring and driving assistance. Precisely, it can not only remove the redundant part in the partial fully connected layer and the network structure but also simplify the backbone network and the repeated feature extraction part, resulting in an obviously improved performance as compared to the current object detection network derived from YOLOV3 [20].

2) *Novel efficient Edge YOLO system based on E-CC*: Different from FL, the respective tasks of the edge and the cloud are more clearly defined in E-CC. The edge carries out model reasoning and data uploading in idle conditions, while the cloud is responsible for the timed training and weight updating of new data independently. In addition, in the most representative AI SOC (NVIDIA Jetson series) [21] edge computing platform, we build a movable platform based on ROS [22], [23] to emulate the object detection procedure in traffic safety monitoring and driving assistance.

3) *Performance evaluation of E-CC solution*: We compare and evaluate the two computing solutions, i.e., FL-based cloud computing and the proposed E-CC computing. It is found that, the Convolutional Neural Networks (CNNs) in the E-CC model is just trained on the cloud and pre-loaded on the movable platform to carry out real-time computation and inference, which makes the proposed solution more efficient and flexible by compared to traditional cloud computing one.

The remaining sections are organized as follows: Section II introduces the related work on object detection with deep learning networks and the development of edge computing devices implemented in vehicles. In Section III, we present the design and algorithm of the Edge YOLO system. In Section IV, we present the results of the experiment and a detailed description. In Section V, we present this study's conclusions and future applications.

## II. Related Work

In this section, we first introduce the application of edge cloud cooperation (E-CC) architecture in ITS and the role of AI in different computing architectures. Then, we review the development history of deep learning algorithms for object detection, including one-stage and two-stage object detection networks and lightweight neural networks [24]. In addition, we describe the embedded computing platform available for ITS.

The future of ITS is considered as the convergence of various new technologies, including AI, IoT, autonomous driving, and so on. The current ITS relies on closed or independent systems, such as the isolation between the single-layer cloud computing model and the edge nodes, which cannot build a more integrated massive ITS. Therefore, based on the E-CC architecture, it is to solve the edge node isolation as well as effective means of transport network management, which alleviate the pressure caused by the mass of data in cloud computing, greatly reducing the calculation time delay. It can be used in IoV, Advanced Driving Assistance System, traffic monitoring scene, and so on.

Extensive attention about AI has been aroused in the field of edge computing recent years. The core issue is how to deploy a series of neural networks, such as CNNs, in the closest location to the user. Therefore, on the one hand, the current research focuses on optimizing the CNNs series of object detection networks to make it more possible to adapt to edge devices. On the other hand, we strive to promote the model reasoning ability in edge devices to acquire a stable balance in edge computing devices between the power consumption and computing power. We believe that AI and edge computing complement each other. AI broadens the application scenarios for edge computing, which in turn provides a heterogeneous platform for the actual application of AI. More importantly, the emergence of AI has re-coordinated the competitive relationship between edge computing and cloud computing and can achieve the complementary advantages of the two.

For object detection, we aim to classify the predefined categories from the target database and determine the coordinate position of the instance in the image, which is the most challenging and practical task in the current computer vision field. Before 2012, the object detection algorithm was mainly based on the traditional computer vision method of geometric representation and manual features. However, since 2012, the Hintion team has broken the record of image classification in the AlexNet Visual Recognition Challenge [25], and successfully applied Rectified Linear Unit (ReLU) in CNNs for the first time. Compared to traditional machine learning algorithms such as Histogram of Oriented Gradient, Support Vector Machine, Hidden Markov Model and so on, convolutional layer can get more pure features from the original RGB channel, thus the application of deep learning in object detection officially opened the prelude. At the same time, the network structure and algorithm based on deep learning have shown their talents in various object detection. The Girshick team first proposed Regions-CNN (R-CNN) [26] in 2014. The candidate window generation method and deep network adopted by it are effective for object detection accuracy. It has greatly improved, and this has also been the most accurate target detector so far. In addition, in order to improve its object detection speed and training efficiency, Fast-RCNN [27] and Faster-RCNN [28] detection networks are derived. Meanwhile, R. Joseph and others proposed the YOLO object detection network in 2015. YOLO is the abbreviation of "You Only Look Once". This is the first single-stage detector in the field of deep learning. Detection paradigm combining 'Proposal and Classifier' of different R-CNN series object detection networks. YOLO uses the entire picture as the network input, directly returns the position and category of the bounding box in the output layer, and obtains the class- specific confidence score of each box.

In the same year, the Single Shot Multibox Detector (SSD) proposed by W. Liu *et al.* [29] introduced multi-scale feature mapping and multi-resolution detection technology, using specific feature maps for detection, and compared with the mean Average Precision (mAP) of another One-Stage detector it has an improvement of 15%-20%. So far, the deep learning scheme in the object detection field has formed two different types: One-Stage object detector is represented by YOLO and SSD, while Two-Stage object is represented by RCNN and Spatial Pyramid Pooling Network (SPPNet). More noteworthily, these two different types of object detection algorithms continue to optimize network structure and learn from each other during the development process. Between them, the network structure of the YOLO series has gone through four generations of evolution: YOLOV1, YOLOV2, YOLOV3, and current YOLOV4. In the process of evolution, an improved methods like joint training algorithms, objectness score, multi-scale prediction, etc. have come into widely utilized. Finally, a network structure is formed where CSPDarknet- 53 is used as Backbone, Path Aggregation Network (PANet) [30] is used as Path-aggregation Neck, SPP-block is used as an additional block, and YOLOV3 is used as Head. YOLOV4 is known as the most perfect detection network that owns the best equilibrium between frames per second(FPS) and Precision so far. Nonetheless, the model reasoning of the YOLOV4 algorithm still needs to be implemented by virtue of desktop-level high-performance GPUs, which cannot be used on a large scale in edge computing scenarios.

Currently, CNNs and other network models are getting more profound and more complex in pursuing higher accuracy. However, in the context of mMTC, this is not applicable to real mobile or embedded computing devices due to computing power and memory limitations. As a result, lightweight neural networks emerge, which is currently achieved through the following three aspects: (1) optimizing the convolution operation and directly designing the small model. For example, group pointwise convolution channel rearrangement and other methods in ShuffleNet [31], MobileNet [32] changes the convolution structure by decomposing a standard convolution factor into a depth-wise conv and a pointwise conv, and completely separate the spatial correlation between the channels. (2) Based on the existing CNNs network through pruning, weight sharing, knowledge distillation model parameters by distillation and optimizing the activation function. (3) Using Neural Architecture Search (NAS) [33] technology in AutoDL [34] to design the Neural Network automatically, the high-efficiency Network Architectures can be obtained under the constraints of model size, number of parameters and other strategies. The latest lightweight network on the mobile terminal is MobileNetV3 [35]. It improves its predecessor in three ways. (1) Introducing A new activation $h-swish$ function to reduce the calculation time and eliminate the potential loss of accuracy. (2) Removing the feature generation layer and use $1 \times 1$ convolution to calculate the feature map. (3) Introducing the squeeze and excitation module to assign weight to feature map.

At present, devices that can implement edge computing include SoC, FPGA, Application Specific Integrated Circuit(ASIC) [36], Central Processing Units (CPU) and GPU. In edge computing, FPGA was once considered the most suitable hardware for AI deployment. Although they have achieved good performance in power consumption control, they are not enough outstanding in reasoning performance and deep learning ecological support. For ASIC, their long development cycle and high cost is suitable for professional customization, but it can hardly adapt to the rapid development of object detection algorithms. Simultaneously, the production process of semiconductor components such as GPU, CPU, and memory develops in full swing, the available edge computing resources are provided for neural networks, which makes it increasingly possible for AI models to be implemented in edge computing. What's more, it is worth mentioning that the NVIDIA Jetson series is a promising AI SoC [37]. It is characterized by high energy efficiency, small size, and high throughput. However, the power consumption of traditional GPU ranges from 100 to 250 watts, while the integrated GPU on Jetson is between 5 to 15 watts. More importantly, it can use migration learning tools to convert the common deep learning frameworks such as Tensorflow, Caffe, and Pytorch into TensorRT [38] to accelerate model inference. The basic principle is to optimize the GPU characteristics, network layer fusion computing, and accuracy during inference. It can provide low-latency, high-throughput deployment reasoning for embedded platforms such as autonomous driving and security monitoring.

## III. System Platform and Algorithm

In this section, we elaborate on our system scheme. The main contents are shown as follows: in section A, we introduce the E-CC system scheme we proposed, including the unit composition and workflow of the system, and introduce the Cloud computing solutions, which is the current mainstream object detection, and make the comparison between them; in section B, we describe the basic framework based on YOLOV4 in detail and propose an improved algorithm Edge YOLO which is adapted to edge computing scenarios. Additionally, we reconstruct the Backbone layer based on the original network, using channel pruning to reduce the network size significantly. We also reasonably reduce the feature fusion techniques that occupy more GPU resources in the original algorithm and have limited improvement to obtain a new architecture that balances accuracy and speed in edge AI scenarios.

### A. System Platform Design and Comparison

In this paper, we have selected the most representative embedded platforms in the Jetson series so far: Jetson Xavier and Jetson nano; this two hardware are widely used in the Jetson series and are uniquely cost-effective. We will use them as autonomous vehicles. The onboard computer system on the platform uses the camera carried by the autonomous vehicle to perform real-time object detection tasks during inspections. Unlike the past model evaluation of high-performance graphics cards on the PC platform, embedded AI computing devices are only used for model reasoning when maintaining low power consumption. Because the model training requires more powerful computing power than the inference process, a dedicated GPU workstation to achieve is needed. We put the model training task on the computing power workstation in the cloud, and the cloud will upload the image from the edge device, and regular training is performed to obtain new weight files and update the model weights of the edge layer while real-time object detection is all implemented by the edge device.

The System Solution shown in Fig. 2 is explained as follows.

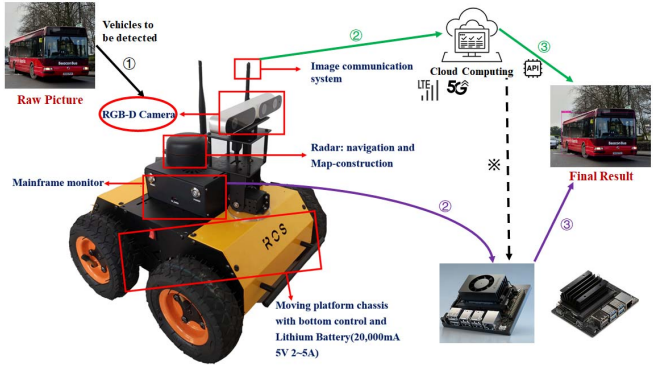

Fig. 2. Two different working paths for object detections based on edge devices: Our E-CC solution (Purple path) and Cloud computing solution (Green path).

Our System Platform (Purple path)

1) The camera (FPS is set to 30) collects the real-time images of the target crowd during the movement of the mobile platform to obtain the original image.

2) The video stream is sent to the edge computing equipment (NVIDIA platform) based on CNNs neural network for model reasoning and object detection.

3) The results will be saved locally. At the same time, the system will feedback the test results.

The cloud server will train the collected data after labeling, and download the updated weight to the edge device.

Cloud computing comparison scheme (Green path)

1) The camera (FPS is set to 30) collects video and transmits it to the central server.

2) The transmitter excites the cloud computing server through API. The cloud server receives the original data of the image, allocates certain computing resources and certain query per second(QPS), and sends it to the object detection network based on CNNs to get the detection results.

3) The cloud server will feedback the results to the local, and the result can be obtained from the edge.

### B. Algorithm and Network Structure

Before performing object detection, we need to extract the necessary feature information from the image and use the features of the image to achieve positioning and classification firstly. The success of deep learning in the image field has a considerable reflection in the feature extraction of the image. This part is called the backbone network. Through this network, we complete feature extraction and directly enter the object classification. However, in Edge YOLO, the task is to perform object detection, and the position of the object in the image cannot be obtained only by relying on the classification network. Therefore, in the object detection network, the backbone network can provide several combinations of receptive field size and center step size, which can satisfy the detection of different scales and categories.

With the mutual reference and integration of structures of various object detection networks, for example, the YOLO series introduces network modules such as Feature Pyramid Networks (FPN) [39] in the RCNN series after YOLOV3, for the detection of different sizes of detection targets, especially small targets, it is necessary to extract information of different scales from different feature maps and performs independent detection on the fused feature maps and further use the feature information extracted by Backbone to complete multi-scale prediction and improve detection performance, and this part is called the Neck network. As shown in Fig. 3, our object detection network Edge YOLO continues the three-in-one framework which consists of feature extraction network, feature enhancement network and object detection network. In the Backbone part, channel pruning is mainly performed to reduce the size of the stack compression model of the residual block. The purpose is to reduce many convolution operations in the entire feature extraction process. In the Neck part, because original network PANet in the YOLOV4 needs to repeatedly extract the feature layer, the bottom-up feature fusion is added based on FPN and then formed a closed loop of the network. Given that repeated extraction of features will cause a lot of redundancy, for lightweight networks that are more suitable for mobile devices or devices without GPU, it often ignores feature fusion and multi-scale prediction. In the Neck layer, we test a large number of feature fusion structures, and finally, the deletion of the PANet network, the combination of SPP and FPN were retained and used to perform feature fusion. Convolutional layers and up-sampling connected two adjacent networks of different scales. Finally, for the YOLO Head part, we still keep the original plan, which is composed of a convolution combination of $3 \times 3$ and $1 \times 1$, and finally linearized output. This part is essentially a regression prediction based on an anchor-based box; In Edge YOLO, we do this with two convolution layers. The Edge YOLO object network is still composed of three sub-networks: Backbone, Neck, and Head. All in all, Edge YOLO is primarily based on (1) The Trimmed Backbone of YOLOV4 (2) Improved Feature Fusion Network in Neck (3) The introduction of activation function Leaky_ReLU (4) Sparse training and model accelerated inference adaptation, these four parts build a complete target detection network.

*1) The Trimmed Backbone of YOLOV4:* The Backbone part of YOLOV4 is CSPDarknet53. For the previous generation of object detection network YOLOV3, the Backbone part is Darknet53, which is composed of the ResBlock [40] stack of the remaining network, and connected by many $3 \times 3$ filters with a step of 2 during the process. However, it requires a large number of samples to calculate, and the detection is extremely time-consuming. So, the basic idea is Skip-connections, which closely connect each network layer. One of the DenseBlock [41] includes multiple convolutional layers; Batch Normalization and Active Function use deeper networks to form a receptive field and finally form DenseNet. YOLOV4 draws on the Cross Stage Partial Network (CSPNet) [42] network structure from the previous generation of YOLOV3 Moreover, it uses the feature map, which is the input of the DenseBlock, to split and divide into two parts, where the main part continues to stack the residuals, and the remaining part is directly connected to the end. The experimental results also prove that although the ResNeXt, which is composed of residual nets (ResNet), has good performance in image classification tasks, its detection accuracy is lower than CSP-Darknet53. This is also an important factor that compared with the mAP of YOLOV3 and YOLOV4 on the COCO2017 datasets can be increased by nearly 10% and the FPS loss is slight. On the other hand, considering that for existing edge computing devices, memory bandwidth and access speed are important factors affecting their performance, CSPDarknet53 will occupy a large amount of memory resources. Fig. 4 shows a fusion strategy. The strategy divides a feature map into two equal parts, which are denoted as Part 1 and Part 2, respectively. Part 1 is directly linked to the end of the stage part. Part 2 will go through a dense block Concatenation. The feature map in Part 1 does not need to go through the Transition Layer composed of Dense different layers in this process. The advantages of DenseNet's feature reuse characteristics are molded in this way. This strategy can improve the learning ability of the network under the premise that the gradient information is almost intact. To reduce the repeated gradients and perform high-performance calculations in Edge YOLO, we use CSPBlock instead of the original ResBlock as the residual network. We aim to reduce the memory usage further, divide the channels in the characteristics layer through the usage of two CSPBlock modules, and finally simplify the residual structure. In this way, CSPBlock can enhance the gradient's difference and improve the convolutional layer's learning ability. In the Backbone part, we will have Convolution, Batch Normalization, and Activate function Leaky_ReLU to form a basic network component CBL and send the four CBLs to the CSPNet network structure for concatenation and finally form the CSPBlock. The Backbone of YOLOV4 is built on CSPDarknet53. Such a feature extraction network requires a large number of network parameters to improve the receptive field, which includes 29 convolutional layers, $725 \times 725$ receptive fields and occupies 88 layers. The feature extraction network also regards Mish as the active function in Backbone. Although Mish has the characteristics of smoothness, non-monotonicity, and smoothness, it has nonlinear characteristics compared to the other two activation functions in the three active function graphs we show in Fig. 5, which makes it more complex. And its occupancy rate of RAM is extremely high. To this end, we introduce the activation function Leaky_ReLU. to solve the Dead Neuron phenomenon when the input value of the original function ReLU is negative. It is a very small constant for the $\alpha$ in Leaky_ReLU, to solve the Dead Neuron phenomenon when the input value of the original function ReLU is negative. It is a tiny constant in Leaky ReLU so that the information on the negative axis will not be lost, and the problem that the ReLU function does not learn when the negative value appears can be solved. The parameter is manually assigned a value through prior knowledge, and then it is used as a parameter for training.

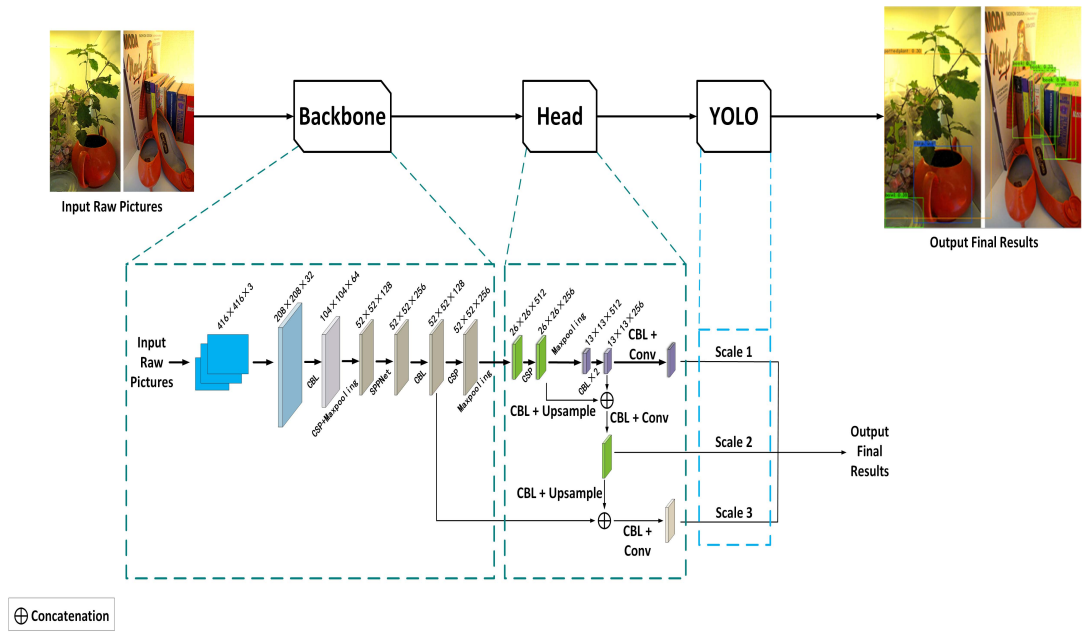

Fig. 3. Overview of the proposed Edge YOLO framework. It contains the improved CSPDarknet, SPPNet, FPN and YOLO heads.

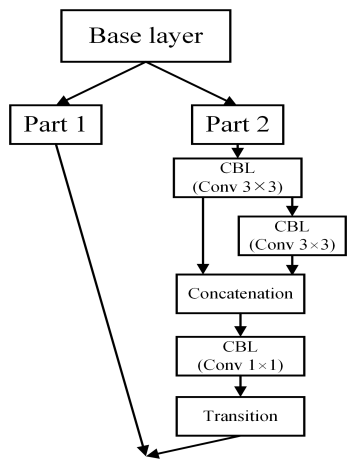

Fig. 4. Detailed block diagram of CSPNet features fusion strategy. In the backbone network of Edge YOLO, the feature map represented by Part 2 is first transited and finally merged with the Feature map represented by Part 1 to effectively reduce the repeated gradient information.

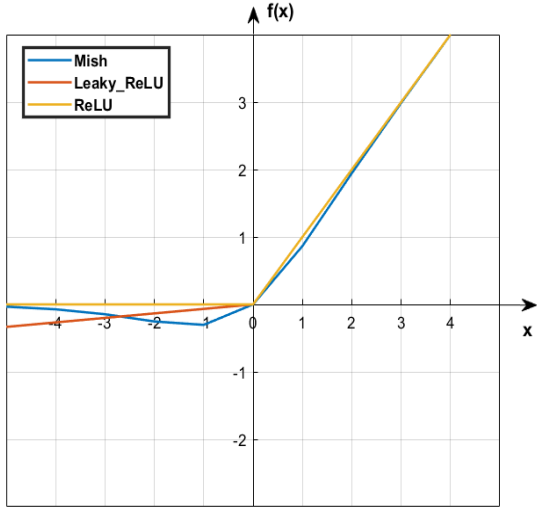

Fig. 5. Comparison of three activation functions: Mish, Leaky_ReLU and ReLU.

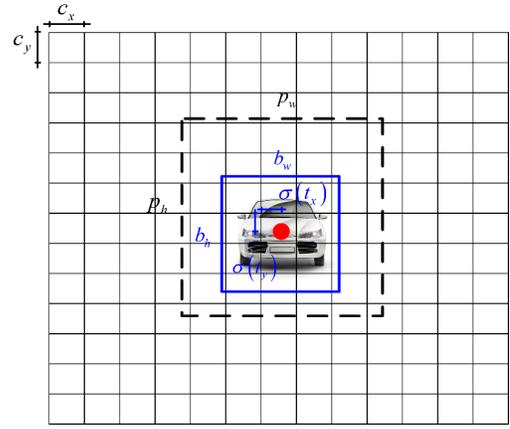

Fig. 6. Bounding boxes with offset and grid location prediction. We predicted the position of the center coordinate of the mask relative to the filter application.

We fix the size of the original input image is to $416 \times 416 \times 3$ (refer to Fig. 3) and introduce the network structure diagram of the entire system, which is simpler than the entire network structure of YOLOV4. The parameter quantity is only about 10% of the original network. A picture is randomly selected from the COCO2017 verification set for detection. Because there is no shoe label in COCO2017, the high heel in the picture cannot be detected. Although it is affected by light and has substantial brightness interference in the background part, Edge YOLO still has a better detection effect in terms of the target.

As presented in Fig. 6, in order to count the most common box sizes from all ground-truth boxes in the training set, we follow the anchor box [43] mechanism in YOLOV3 to optimize the range of predicted objects and combine them with the prediction of bounding box and add its prior experience for multi-scale prediction. For Edge YOLO, the center position of the target object falls on the center position of the $13 \times 13$ cell, the length and width of each cell's coordinates are $c_x$, $c_y$, and the size range of each grid is $1 \times 1$. The center coordinates predicted by bounding box are $b_x$, $b_y$ and width and height are $t_w$, $t_h$ respectively. The absolute center coordinates of the final output are compressed by the Sigmoid function to ensure that the target center is in the grid unit and reduces the offset. For the anchor box width and height are $p_w$, $p_h$ respectively. The details are as follows:

$$\begin{aligned} b_x &= \sigma(t_x) + c_x \\ b_y &= \sigma(t_y) + c_y \\ b_w &= p_w \times e^{t_w} \\ b_h &= p_h \times e^{t_h} \end{aligned} \quad (1)$$

*2) Improved Feature Fusion Network in Neck:* The Neck structure is to extract the fusion features of the Backbone

**Algorithm 1** Generate Anchor Box Through K-Means Clustering

**Input:** $D = \{(x_1, h_1), (x_2, h_2)\ldots, (x_m, h_m)\}$, $K$
    $D$ is the list of Box objects of bounding Boxes
    $K$ Number of clusters
**Output:** Optimized Anchor cluster division
    $P = \{C_1, C_2\ldots, C_k\}$.

1: **function** ANCHOR_K-MEANS(boxes, n_anchors, centroids)
2:     $C \leftarrow \{\}$
3:     $D$ Normalization
4:     Randomly initialize $\mu_i$ $(i = 1, 2, \ldots k)$
5:     **for** $j = 1, 2 \ldots, k$ **do**
6:        $d_{ji} = \|x_j - \mu_i\|^2$
7:        $v_j = \arg\min_{i \in \{1,2,\ldots,k\}} d_{ji}$
8:        $C_{v_j} = C_{v_j} \cup \{x_j\}$
9:     **end for**
10:    **for** $i = 1, 2 \ldots, k$ **do**
11:       $\mu'_i = \frac{1}{|c_i|} \sum_{x \in C_i} x$
12:       **if** $\mu_i \neq u'_i$ **then**
13:          $\mu_i \leftarrow u'_i$
14:       **end if**
15:    **end for**
16:    when current vector $u_i$ is not updated
17: **end function**

output better. At present, almost all the evolution of the object detection field comes from the tricks in Neck. In the Edge YOLO part, we choose the fusion of SPP+FPN modules.

SPPNet, which module was first used in the two-stage object detection network of the RCNN series. Unlike the ordinary CNN structure, the input image size needs to be fixed. The cropping operation will cause the content of the required object to be lost, making it possible to produce a fixed size output for any size. In the YOLOV4 series, the network is greatly tailored, only the four largest pooling parts are retained, and there is no need to use a fully connected layer to achieve it, but this also facilitates our multi-scale training to improve the accuracy of small object detection. In Fig. 7, the scale division of the multi-scale feature map is presented. Our SPP Block continues using the maximum pooling method used by YOLOV4: $k = \{1 \times 1, 5 \times 5, 9 \times 9, 13 \times 13\}$ to output the feature map and then perform concatenate combination. This combination and pooling method increase the acceptance range of the feature backbone feature that separates the context, thus enhancing the receptive field.

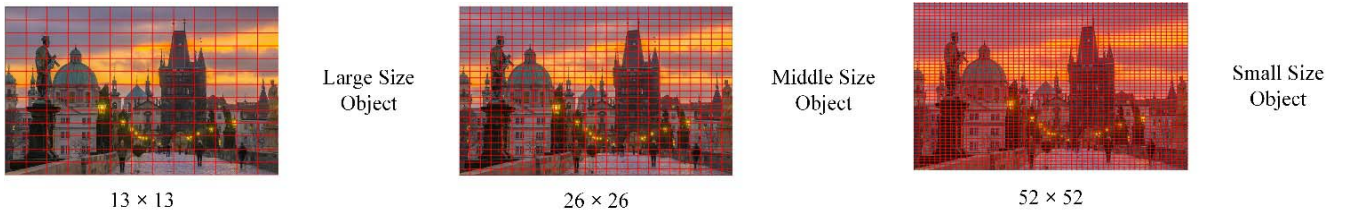

Fig. 7. Edge YOLO has made several multi-scale predictions, including large, middle, and small size. In Edge YOLO, aiming at the three different object detection sizes above, we make the input picture size is 416 × 416, and the output sizes are 13 × 13, 26 × 26, 52 × 52.

FPN refers to a scheme of building a feature pyramid inside the convolution, which is essentially a feature detector. For object detection, the top-level features are mostly used to predict. However, in Edge YOLO, FPN is introduced to make a comparison under the premise of less calculation, and the feature maps with strong semantic information at low resolution are integrated with feature maps with weak semantic information at high resolution to finish multi-scale prediction truly. But unlike PANet in YOLOV4, PAN is deleted partly because there are many repeated feature extraction processes. In Fig. 8 three feature maps are still extracted, the sizes of which are 13 × 13, 26 × 26, 52 × 52. FPN includes two paths, bottom-up and top-down. We use down-sampling in the bottom-up path and up-sampling in the top-down path, and the step size is two. In this process, the spatial resolution is reduced, but the semantics of each layer is increased and can be more accurately activated locally. These features will be horizontally connected and enhanced by the bottom-up path. They use a bottom-up network structure with lateral connection to construct feature fusions of different sizes with high-level semantics. The rough resolution map relates to 1 × 1 convolution through double up-sampling to finally generate the finest resolution map. This method can enrich the high-level semantics of each layer while ensuring a certain speed and without sacrificing the ability of representation. It is undoubtedly very cost-effective for a lightweight network to achieve small object detection.

The object detector is executed in the form of a sliding window, so there are multiple detection anchors (18 anchors are selected in this article) corresponding to the same target object. Non-Maximum Suppression (NMS) removes redundant frames and finds the most matched results. It calculates the list $B$ of Bounding Boxes, the corresponding confidence level $C$ and sets the threshold $T_{nms}$. By selecting the detection frame with the maximum score and comparing the remaining detection frames with the IoU of the maximum score detection frame, it continuously eliminates the boxes more significant than the threshold until the list $B$ is empty. However, this algorithm directly deletes the lower confidence level when the overlap rate of the two target frames is high. Generally, this algorithm can lead to missed detection and a low recall rate. We use an optimization algorithm called Soft-NMS [44] to adjust the score of the score and output the result through the function of the IoU, instead of directly setting the score to zero.

We have followed the three-loss function calculation methods for the Loss function since YOLOV3. They are object localization offset loss1: $loss_1$, object confidence loss: $loss_2$, and object classification loss: $loss_3$, respectively In object classification loss, we use the binary cross-entropy loss for summation to make up for the shortcomings of gradient dispersion. Among them, $I_i^{obj}$ represents whether there is an object in the predicted object frame, While A represents the predicted bounding box and B represents the true bounding box:

$$loss_{\text{total}} = loss_1 + loss_2 + loss_3 \quad (2)$$

1) object localization offset loss:

$$loss_1 = \sum_{i=0}^{S^2} I_i^{obj} \sum_{c \in classes} (p_i(c) - \hat{p}_i(c))^2$$
$$+ \sum_{i=0}^{S^2} I_i^{obj} \sum_{c \in classes} (\log p_i(u) - \log \hat{p}_i(u))^2 \quad (3)$$

2) object confidence loss:

$$loss_2 = \sum_{i=0}^{S^2} \sum_{j=0}^{B} I_{i,j}^{obj}(C_i - \hat{C}_i)^2 + \lambda_{noobj} \sum_{i=0}^{S^2} \sum_{j=0}^{B} I_{i,j}^{noobj}(C_i - \hat{C}_i)^2$$
$$(4)$$

3) Unlike, the Mean-Square Error (MSE) algorithm used by YOLOV3 to calculate the classification loss, we extend the Complete Intersection Over Union (CIoU) algorithm adopted by YOLOV4 to calculate the object localization offset loss.

$$loss_3 = 1 - IoU(A, B) + \frac{\rho^2(b, b^{gt})}{c^2} + \alpha v \quad (5)$$

$$v = \frac{4}{\pi^2} \left( \arctan \frac{w^{gt}}{h^{gt}} - \arctan \frac{w}{h} \right)^2 \quad (6)$$

$$\alpha = \frac{v}{(1 - IoU) + v} \quad (7)$$

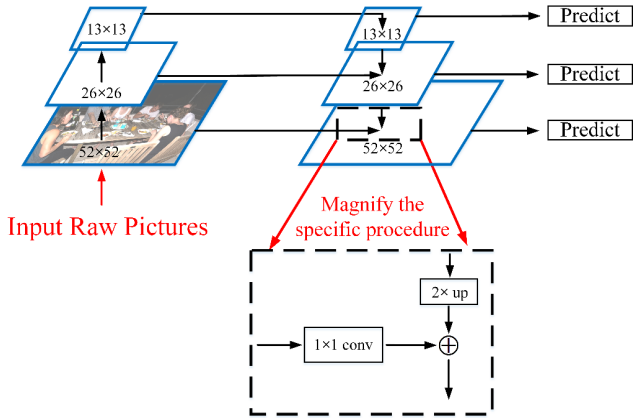

Fig. 8. FPN applied in the one-stage architecture. Build a module by combining horizontal joins and top-down paths by addition.

We use the KITTI and COCO datasets for training and obtain the above result graphs. In Fig. 9, we train the network for 100,000 Batches. In terms of the optimizer, we improve the efficiency of regular training with stochastic gradient descent (SGD). Compared with Batch gradient descent to calculate the gradient with all data, SGD only performs the update once to obtain the corresponding parameters each time in the new training set, which has a faster training speed and supports the addition of the samples can be added. After the repeated adjustments, we set $\eta$ as $= 0.0002$.

$$\theta = \theta - \eta \cdot \nabla_\theta J(\theta; x^{(i)}; y^{(i)}) \tag{8}$$

**Algorithm 2** Soft-NMS Method
**Input:** $B = \{b_1, \ldots b_N\}$, $C = \{c_1, \ldots c_N\}$, $T_{nms}$
  $B$ refers to is the list of initial detection boxes
  $C$ refers to is the collection of corresponding detection scores
  $T_{nms}$ refers to is the NMS threshold
**Output:** List of final detection Bounding boxes $S$

1: **function** SOFT_NMS(boxes, confidences, nms_threshold)
2:   $S \leftarrow \{\}$
3:   **while** $B \neq empty$ **do**
4:     $m \leftarrow \arg \max C$
5:     $S \leftarrow S \cup b_m$; $B \leftarrow B \cup b_m$; $C \leftarrow C \cup c_m$
6:     $s_i = s_i e^{-\frac{IoU(M, b_i)}{\sigma}}$
7:     **for** $b_i \in B$ **do**
8:       **if** $IoU(M, b_i) \geqslant T_{nms}$ **then**
9:         $B \leftarrow B \cup b_i$; $C \leftarrow C \cup c_i$
10:      **else**
11:         $s_i \leftarrow s_i f(IoU(M, b_i))$
12:      **end if**
13:    **end for**
14:  **end while**
15: **end function**

In TABLE I, the filter, stride and Billion Floating-Point Operations Per Second (BFLOPS) of each layer are shown

TABLE I
IMPROVED BACKBONE OF EDGE YOLO FOR EACH LAYER

| Layer | Type | Size/Stride | Filters | Output | BFLOPS |
|---|---|---|---|---|---|
| 0 | Conv | $3 \times 3/2$ | 32 | $208 \times 208 \times 32$ | 0.075 |
| 1 | Conv | $3 \times 3/2$ | 64 | $104 \times 104 \times 64$ | 0.399 |
| 2 | Conv | $3 \times 3/2$ | 64 | $104 \times 104 \times 64$ | 0.797 |
| 3 | Route | | 2 | $104 \times 104 \times 32$ | |
| 4 | Conv | $3 \times 3/1$ | 32 | $104 \times 104 \times 32$ | 0.199 |
| 5 | Conv | $3 \times 3/1$ | 32 | $104 \times 104 \times 32$ | 0.199 |
| 6 | Route | | 54 | $104 \times 104 \times 64$ | |
| 7 | Conv | $1 \times 1/1$ | 64 | $104 \times 104 \times 64$ | 0.089 |
| 8 | Route | | 27 | $104 \times 104 \times 128$ | |
| 9 | Max | $2 \times 2/2$ | | $52 \times 52 \times 128$ | 0.001 |
| 10 | Conv | $1 \times 1/1$ | 128 | $52 \times 52 \times 128$ | 0.089 |
| 11 | Max | $5 \times 5/1$ | | $52 \times 52 \times 128$ | 0.009 |
| 12 | Route | | 10 | $52 \times 52 \times 128$ | |
| 13 | Max | $9 \times 9/1$ | | $52 \times 52 \times 128$ | 0.028 |
| 14 | Route | | 10 | $52 \times 52 \times 128$ | |
| 15 | Max | $13 \times 13/1$ | | $52 \times 52 \times 128$ | 0.058 |
| 16 | Route | | 15 13 11 10 | $52 \times 52 \times 512$ | |

TABLE II
NETWORK ARCHITECTURE OF IMPROVED FPN FOR EACH LAYER

| Layer | Type | Size/Stride | Filters | Output | BFLOPS |
|---|---|---|---|---|---|
| 17 | Conv | $1 \times 1/1$ | 256 | $52 \times 52 \times 256$ | 0.709 |
| 18 | Conv | $3 \times 3/1$ | 128 | $52 \times 52 \times 128$ | 1.595 |
| 19 | Route | | 18 | $52 \times 52 \times 64$ | |
| 20 | Conv | $3 \times 3/1$ | 64 | $52 \times 52 \times 64$ | 0.199 |
| 21 | Conv | $3 \times 3/1$ | 64 | $52 \times 52 \times 64$ | 0.199 |
| 22 | Route | | 21 20 | $52 \times 52 \times 128$ | |
| 23 | Conv | $1 \times 1/1$ | 128 | $52 \times 52 \times 128$ | 0.089 |
| 24 | Route | | 18 23 | $52 \times 52 \times 256$ | |
| 25 | Max | $2 \times 2/2$ | | $26 \times 26 \times 512$ | 0.001 |
| 26 | Conv | $1 \times 1/1$ | 128 | $26 \times 26 \times 128$ | 0.044 |
| 27 | Conv | $3 \times 3/1$ | 256 | $26 \times 26 \times 256$ | 0.399 |
| 28 | Route | | 27 | $26 \times 26 \times 128$ | |
| 29 | Conv | $3 \times 3/1$ | 128 | $26 \times 26 \times 128$ | 0.199 |
| 30 | Conv | $3 \times 3/1$ | 128 | $26 \times 26 \times 128$ | 0.199 |
| 31 | Route | | 30 29 | $26 \times 26 \times 256$ | |
| 32 | Conv | $1 \times 1/1$ | 256 | $26 \times 26 \times 21$ | 0.089 |
| 33 | Route | | 32 27 | $26 \times 26 \times 512$ | |
| 34 | Max | $2 \times 2/2$ | | $13 \times 13 \times 512$ | 0.000 |

in the Backbone part. Compared with the Backbone network structure part of YOLOV4, the number of layers is reduced to 17 from 88. And only 6 convolutional layers are used for feature extraction. In TABLE II FPN is reintroduced to replace PANet in the Neck layer. In the Neck part of YOLOV4, 20 convolutional layers are used and up-sampling, down-sampling are performed twice, thereby using feature fusion and making the extracted region of interest (ROI) [45] feature richer, 20 convolutional layers are used and up-sampling, down-sampling are performed twice. Edge YOLO cuts down the bottom-up path augmentation, Adaptive feature pooling, Fully-connected fusion and other parts of PANet. On this basis, the number of layers in TABLE II are reduced to 17 from 61. from 61 to 17.

## IV. RESULTS AND DISCUSSION

In this section, the dataset, construction, and training process of the training cloud platform, as well as the inference process of the model on the edge computing device, are elaborated. In addition, the results are evaluated. The finally trained model is used in COCO2017 standard datasets to verify and compare it with the two standardized object detection networks (i.e., YOLOV3 and YOLOV4) and two mainstream lightweight object detection networks (i.e., YOLOV3-Tiny and MobileNetV3 SSD). Among them, the threshold of IoU is set to 0.5. If the IoU value is greater than 0.5, it will be judged as True Positive, vice versa. Finally, the mAP, sizes, BFLOPS, and the corresponding FPS of two edge computing devices during autonomous vehicle inspection are measured. Additionally, the COCO dataset created by us is selected. This dataset will regularly add the data collected by edge devices, which can be used for mask detection. The online data annotations will be performed in the backend to retrain and update the weight. The real-time detection of masks is carried out by using the foregoing five object detection networks and the real edge scene of the two NVIDIA Jetson autonomous vehicle platforms built by us. Finally, our solution is compared with the traditional cloud computing solution (where the cloud provides an online software development kit (SDK) [46] for the edge). In addition, the actual real results of the two computing modes are obtained, the results are evaluated through the following experiments, and the detection results are displayed.

### A. Dataset Description

Firstly, COCO2017 is chosen as the dataset, which is a public dataset originating from the Common Objects in Context (COCO) [47] datasets that Microsoft funded and labeled in 2014. The COCO competition is considered the most authoritative benchmark in the field of object detection. There are eight types of annotations, including object instances, object key points, and image captions. Among them, the number of targets included in each image is twice to three times that of other mage datasets, which are widely used as benchmark data in object detection, image segmentation, and image classification. This dataset covers 12 major categories of targets, 80 subcategory targets, and 118278 pictures, where the mAP value of each target is the average value of each type detection accuracy, and the IoU threshold is set to 0.5.

In addition, the KITTI dataset - the most representative dataset in ITS is selected, which is used to measure stereo, optical flow, visual odometry, and the performance of computer vision technology such as Object Detection and Tracking in the vehicle environment. The data set of the target detection part is selected. In this section, there are 7,481 training datasets and 7,518 test sets, including $'Car'$, $'Van'$, $'Truck'$, $'Pedestrian'$, $'Person\_sitting'$, $'Cyclist'$, $'Tram'$, $'Misc'$ or $'DontCare'$ totaling 9 types of targets. Meanwhile, the datasets are taken from two color cameras from the perspective of the driver. The datasets include multi-scale targets such as nearby pedestrians and trams, and distant small vehicles. Each image is described with the degree of truncation or occlusion. The image resolution is $1242 \times 375$, which is saved in jpg format. In TABLE III, the training set, test datasets, and boxes per image for two types of datasets are introduced.

### B. Model Training

In the training process, we work in the Ubuntu 18.04 system environment. When training MobileNetV3 SSD in software,

TABLE III
DESCRIPTION OF COCO2017 AND KITTI OBJECT DATASETS

| Dataset | COCO2017 | KITTI 2D Object Datasets |
|---|---|---|
| Classes | 80 | 9 |
| Training dataset | 106450 | 7481 |
| Test dataset | 11828 | 7518 |
| Boxes of per image | 5 | 15.2 |

TABLE IV
MODEL TRAINING AND RESULTS

| Computer Platform | GPU | Memory | GFLOPS (FP16) | Thermal Design Power | Manufacturing Process | Price |
|---|---|---|---|---|---|---|
| RTX 2080Ti | Nvidia PASCAL 4352 CUDA | 11GB 35 GDDR5X | 26900 | 250W | 12nm | $999 |
| Jetson Nano | Nvidia Maxwell 4395 CUDA | 4GB 64bit LPDDR4 | 472 | 10W | 20nm | $299 |
| Jetson Xavier Nx | Nvidia Volta 4395 CUDA | 8GB 256-Bit LPDDR4x | 2115 | 15w | 12nm | $99 |

Pytorch 1.6 is used as the framework, while Darknet [48] is used as the training framework, which introduces the CUDA 10.1 to accelerate the process of model training while CUDA10.2 accelerate the model inference of Jetson series edge computing devices when the four types of object detection networks in the YOLO series are being trained. And the deep neural network acceleration library cuDNN is 7.6.4 and 7.6.5, respectively. Among them, in terms of hardware, in TABLE IV we give a detailed list of hardware configurations including GPU, Memory, Giga Floating-Point Operations Per Second(GFLOPS), Thermal Design Power and Manufacturing Process. In addition, the detailed hardware environment of our deep learning workstation in the cloud: CPU is Intel Xeon Silver 4210@ 2.20GHz, GPU RTX2080Ti, and the Memory is 128GB of LPDDR4 RAM. The computing power of edge computing devices is only 1.7% and 7.8% for cloud servers, while the power consumption is only 4% and 6%.

As shown in Fig. 9, Edge YOLO and four other representative object detection networks are trained on COCO2017 datasets and COCO datasets, respectively. Among them, YOLOV4 and YOLOV3 object detection networks have been trained for nearly 150,000 Batches to achieve convergence. Furthermore, other object detection networks have been trained for nearly 100,000 Batches. We show the relationship curve between the last 10,000 Batches and Loss. Edge YOLO owns a relatively simple network model. DIoU [49] is introduced to calculate Loss, and the speed of convergence is faster. It can be seen from the figure that a good training effect has been achieved.COCO2017 datasets detect more objects and has many parameters, so the learning rate and parameters need to be adjusted multiple times. To improve the training speed and use the binary cross-entropy to calculate the loss function, the gradient strength decreases with the increase of confidence level and fluctuates up and down when decreasing.

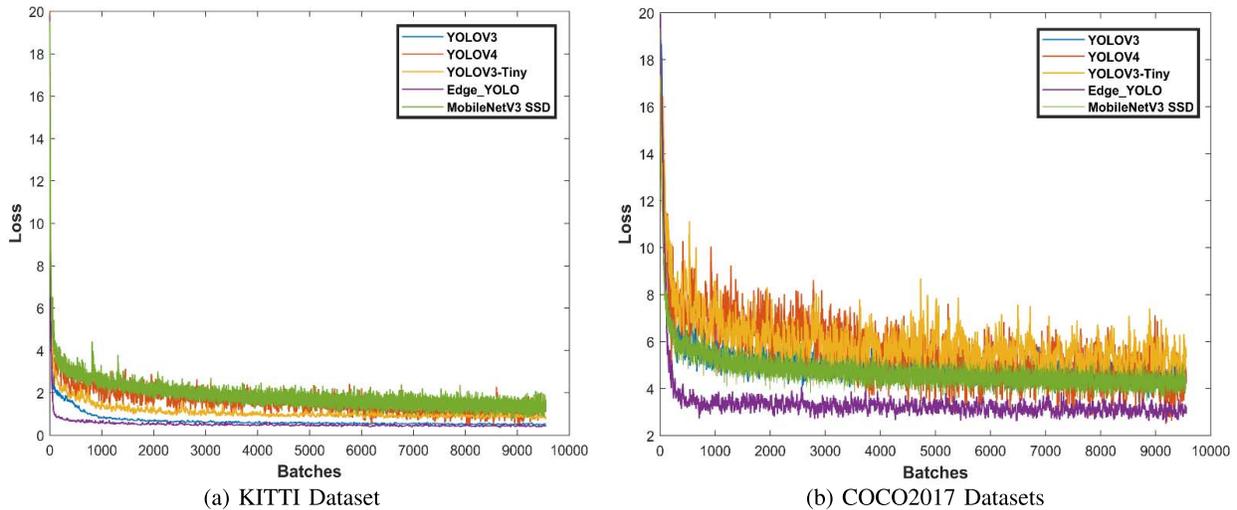

Fig. 9. Training results for loss function by two Datasets.

TABLE V
COMPARISON OF OUR METHOD AND OTHER TARGET DETECTION NETWORK RESULTS ON THE COCO2017 DATASETS

| Model | Sizes(MB) | mAP@0.5 | $FPS^{ave}_{Xavier}$ | $FPS^{ave}_{Nano}$ | FLOP/S | mAP@0.5 (TensorRT) | $FPS^{ave}_{Xavier}$ (TensorRT) | $FPS^{ave}_{Nano}$ (TensorRT) |
|---|---|---|---|---|---|---|---|---|
| YOLOV3 | 236.52 | 0.553 | 4.9 | 2.0 | 65.879G | 0.373 | 31.2 | 4.93 |
| YOLOV4 | 245.78 | 0.657 | 2.1 | 1.0 | 128.459G | 0.459 | 26.1 | 4.62 |
| MobileNetV3 SSD | 115.09 | 0.362 | 10.3 | 9.2 | 17.863G | 0.248 | 32.9 | 20.9 |
| YOLOV3-Tiny | 33.79 | 0.331 | 50.3 | 15.2 | 5.571G | 0.077 | 67.5 | 35.8 |
| Edge YOLO | 25.27 | 0.473 | 26.6 | 11.4 | 10.250G | 0.325 | 49.5 | 26.6 |

## C. Model Performance Analysis

The evaluation of object detection is mainly evaluated by Precision, Recall, FPS, and mAP. For object detection results, it is divided into four categories: TP represents the positive sample of the correct classification, FP represents the positive sample of the misclassification, FN represents the negative sample of the misclassification, and TN represents the negative sample of the correct classification. Precision is the ratio of the samples that are correctly classified as positive examples to the total number of detection samples. The recall is the ratio of samples of this type that are correctly classified and identified to the number of samples in the total target test sets of this type. Moreover, the missed detection rate is used to identify and evaluate the performance of the algorithm for the target frame is formulated as follows:

$$Re = \frac{TP}{TP + FN}$$
$$Pr = \frac{TP}{TP + FP} \quad (9)$$

We attempt to obtain all possible values about Precision and Recall and get the Precision-Recall curve by calculating. The average precision of the network is computed as the area under curve and the value ranges from 0 to 1:

$$AP = \int_0^1 p(r)\,dr \quad (10)$$

In TABLE V, 90% of the COCO datasets is used as the training set. Among the validation set, IoU is set as 0.5, and the input image size is compressed to $416 \times 416 \times 3$.

Compared with the other four models, the size of our algorithm Edge YOLO is only 25.27MB, which is only 10.2% of the original network structure YOLOV4. However, its mAP is up to 47.3%. For the test part of the datasets, files shall be submitted to the MS COCO competition platform to obtain the mAP value. Even in the Jetson Nano with weaker performance, the inference speed FPS of Edge YOLO can reach 11.4. Comparatively, there is a perfect balance between accuracy and precision. The results in the table reflect that model inference for edge devices requires slight processing of the original network model. Otherwise, the real-time requirements will not be met. Practically, hardware acceleration tools for inference acceleration are commonly used in target detection networks tend. Hence, we choose the most representative NVIDIA deep learning engine TensorRT for network calculation merger (inference accuracy is FP16). However, as the data type changes to FP16 during the accelerated reasoning process, the mAP inevitably drops to a certain extent for the very large COCO2017 datasets.

In the experiment, the backbone network MobileNetV3 is applied to the detection network SSD. The official test result for the COCO test sets shows that the mAP of MobileNetV3-SSD increased to 22.0, which increased by 3.2, and the operating latency decreased to 173ms.

Fig. 10 shows the results of object detection when the object is cluttered, the light is strong, or there is object occlusion. For mobile platforms, persons, vehicles, animals, etc., are the most important detection targets. The picture shows that Edge YOLO can accurately detect different objects such as persons, buses, bicycles, and other targets in chaotic scenarios, different distances, and conditions. Simultaneously, pictures (f), (h) detect multiple objects in one picture in a complex environment. Hence, higher accuracy can hardly be guaranteed.

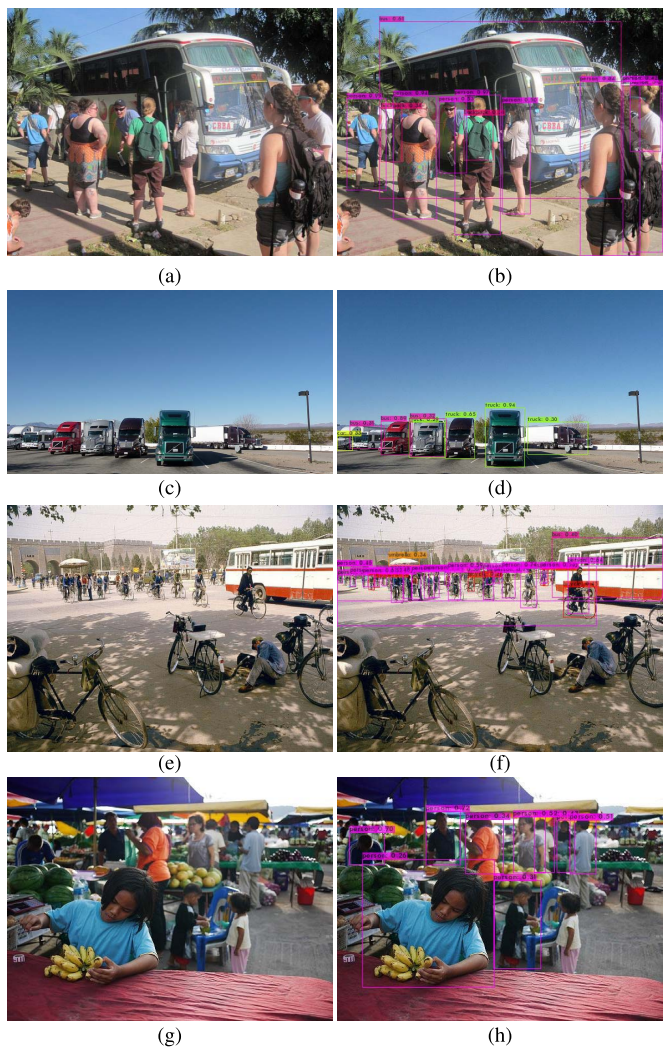

Fig. 10. Test examples for common objects in COCO dataset detected by Edge YOLO. On the left is the original image, and on the right is the detected image.

We choose the sample pictures from COCO2017 Datasets, including persons, carriers, articles, and other small objects.

For edge model inference, its power consumption is the most concerned issue to ensure its effectiveness. Fig. 11 shows the results of the interference of Edge YOLO after training on the KITTI and COCO2017 datasets. Jetson Xavier Nx and Jetson Nano are selected as the vehicle embedded platforms for model reasoning. When the mobile platform performs model inference in real scenarios, the functions such as path planning, obstacle avoidance, and navigation are mainly realized by the technologies including onboard radar and inertial navigation, instead of relying on the camera. For the execution of the Edge YOLO object detection algorithm throughout the process, Jetson Xavier and Jetson Nano spent 5s and 10s loading the model, respectively. When the model is loaded, the power value shows a surge increase. When it is maximized, the power value of the entire edge computing device tends to be stable. Importantly, the optimized feature fusion network of Edge YOLO effectively alleviates the sensitivity of embedded devices to memory resources while maintaining high detection accuracy. For MobileNetV3 SSD, the GPU and CPU are applied in model inference in the lightweight network. As a result, the upper limit of power consumption is lower than that of the YOLO series object detection network. For COCO2017 datasets, its average model loading speed is more obvious because the target number of weight files and data in the set is much lower than that of COCO2017 datasets. (The average loading speed in Jetson Xavier is 3∼5s faster than that of COCO2017 datasets, and the average loading speed in Jetson Nano is 5s∼8s faster than that of COCO2017 datasets.)

In addition, KITTI training datasets are used to retrain the above five network frameworks. The reason lies in that the number of training sets in this testing set is only12% of that of COCO2017 datasets, and the overall target category is 10% of COCO2017. The results in Table VI show reveal that it is moderately difficult to detect the datasets. Therefore, the mAP value is [0.7904, 0.8473]. Compared with COCO2017, the accuracy degree has been improved significantly. At the same time, the final scale and computation amount of the network model is lowered slightly, and the reasoning speed is improved mildly. Obviously, the compression of the trunk layer and the reconstruction of the neck part can considerably reduce the number of Edge YOLO parameters. This can better meet the real-time requirements and realize a fast multi-object detection for the edge scene of vehicles. However, in the scenario of super multi-target detection, it is slightly inferior to YOLOV4 due to the limited number of network layers and parameters.

The detection effects of the Edge YOLO, YOLOV4, and YOLOV3-Tiny object detection algorithms are shown in Fig. 12 respectively. Comparatively, Edge YOLO still maintains favorable detection robustness, although the target is incomplete or blocked. Moreover, the image with high resolution has a positive impact on the detection performance of small-scale targets.

### D. Comparation With Cloud Computing

So far, there have been few edge computing devices that can directly deploy deep learning algorithms. In most cases, we need to connect this kind of device to the cloud computing center (provide SDK by means of cloud server and then achieve cloud access) or carry out local real-time object detection or build a small computer room locally with a fog computing solution. In this experiment, we simulate the real working environment of cloud computing applied in ITS and deployed the proposed Edge YOLO in the cloud computing server, where the cloud server environment is consistent with the deep learning workstation in the experiment. The computing node at the edge obtains the detection request from the cloud server in the form of SDK. We deploy Edge YOLO on cloud servers, and Jetson Xavier, we parallelly upload the image data collected by the edge device to the cloud server through the LTE network. And our network environment is as follows: Upstream bandwidth: 61.4mbps, Downlink bandwidth = 20.35mbps, Ping = 14ms, and Packet loss rate = 0%. In Fig. 13, Pictures (a), (b), (c) are the actual results of a cloud server deployment of Edge YOLO. Picture (d), (e), and (f) indicate the detection results of edge computing equipment after completing inference acceleration due to lower accuracy loss, as can be seen in Fig. 14. Edge YOLO is deployed on cloud computing workstations (with GTX 2080Ti graphics) and on the Edge (Jetson Xavier Nx). The high-performance computing card shows high reasoning ability. However, the cloud computing solution requires testing the transmission delay, which has been proven to slow down processing at the edge. The computing workstation is configured with a 60MB dedicated broadband network. Eventually, the transmission pressure caused by cloud computing speeds up the delay as the number of images increases. Besides, real-time and fast performance is hardly guaranteed. Satisfactory computing performance is shown, regardless of the original Edge YOLO network, the backward and forward reasoning networks accelerated by TensorRT inference. This demonstrates that the edge-cloud collaboration architecture is advantageous in deep edge learning and gives full play to the timeliness of the edge.

TABLE VI
RESULTS FOR KITTI 2D OBJECT DATASET

| Model | Sizes(M) | mAP | $FPS^{ave}_{Xavier}$ | $FPS^{ave}_{Nano}$ | BFLOP/S | mAP@0.5 (TensorRT) | $FPS^{ave}_{Xavier}$ (TensorRT) | $FPS^{ave}_{Nano}$ (TensorRT) |
|---|---|---|---|---|---|---|---|---|
| YOLOV3 | 234.94 | 0.822 | 11.8 | 2.1 | 65.319G | 0.731 | 29.8 | 5.03 |
| YOLOV4 | 244.20 | 0.8473 | 5.2 | 5.1 | 127.365G | 0.784 | 15.5 | 5.72 |
| MobileNetV3 SSD | 33.86 | 0.718 | 11.4 | 6.1 | 12.521G | 0.653 | 56.3 | 26.4 |
| YOLOV3-Tiny | 33.11 | 0.7804 | 62.7 | 15.4 | 6.811G | 0.704 | 66 | 36.6 |
| Edge YOLO | 24.48 | 0.8212 | 40.6 | 9.7 | 9.970G | 0.726 | 68.2 | 25.5 |

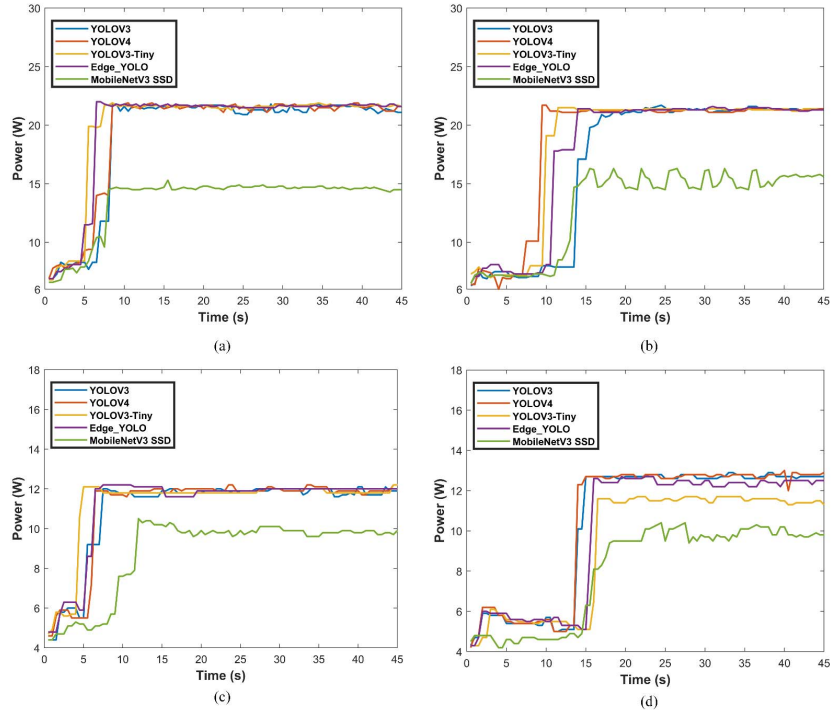

Fig. 11. During the operation of the vehicle-mounted device, we apply five different object detection networks and choose a fragment from the integral energy consumption. Based on two different datasets and two embedded platforms, we divide the model reasoning into four types: namely (a) COCO2017 datasets with Jetson Xavier, (b) COCO2017 datasets with Jetson Xavier, (c) COCO2017 datasets with Nano, and (d) COCO2017 datasets with Nano.

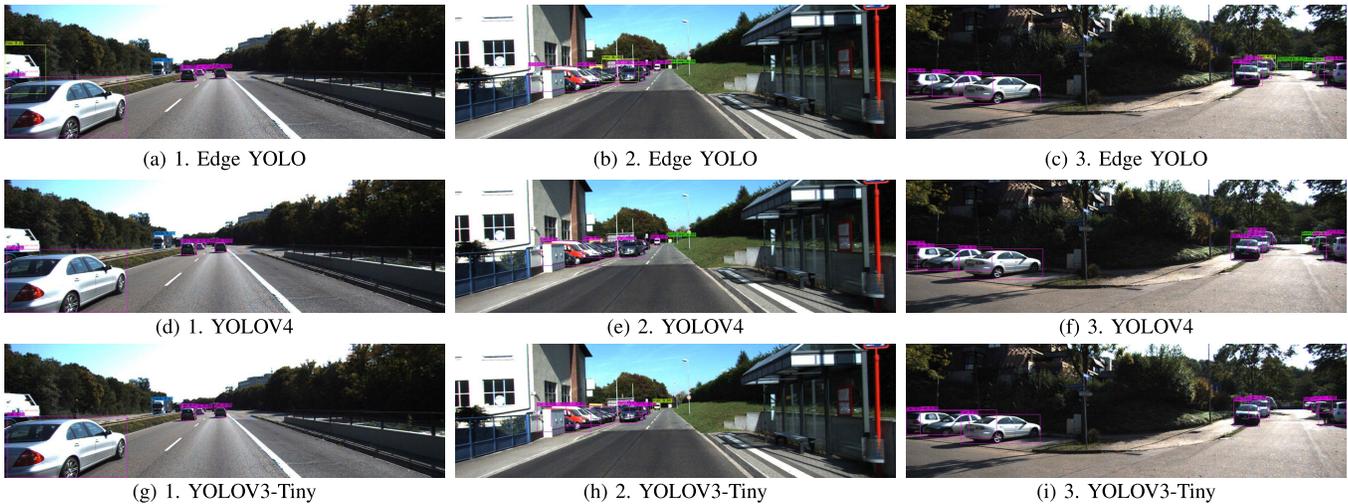

Fig. 12. The results of KITTI 2D detection test set based on Edge YOLO, YOLOV4 and YOLOV3-Tiny models. Edge YOLO shows a high generalization performance, which improves the detectability of small targets and alleviate the risk of missed or false detection.

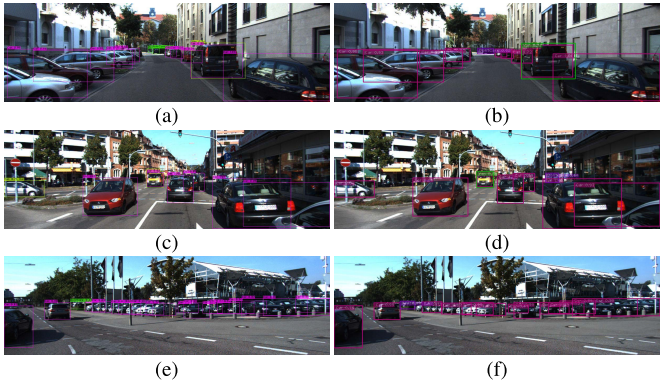

(a) (b)
(c) (d)
(e) (f)

Fig. 13. Test results from KITTI 2D detection test set detected by Edge YOLO and the same dataset trained by cloud computing. Where (a), (c) and (e) represent the detection effect of the original Edge YOLO network located in the cloud. Meanwhile, (b), (d) and (f) represent the detection effect of Edge YOLO at the Edge end after TensorRT accelerated reasoning.

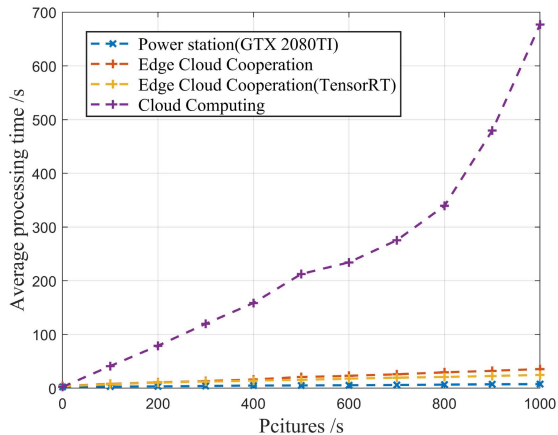

Fig. 14. Number of uploaded pictures and delays calculated by three different approaches: Cloud computing, Edge-Cloud computing (Jetson Nano) and Edge-Cloud computing (Jetson Xavier Nx).

## V. CONCLUSION

In the present study, an E-CC-based movable object detection system, Edge YOLO, is proposed, which is suitable for edge computing devices. This system can not only excellently support the object detection algorithm based on deep learning in the edge computing scenario but also provide a particular methodology for AIoT integrative development. Notwithstanding the limited computing power and pony-size edge devices, Edge YOLO can still guarantee high accuracy. Unlike an offline system, it uploads the data collected at the edge and continuously updates the entire network model assisted by the E-CC. Simultaneously, the edge computing devices equipped with this system present more robust real-time detectability. In Edge YOLO, there are only 8 million parameters in the entire network, which fits the small-scale embedded devices with a single GPU perfectly. In this way, a better balance between the speed and accuracy of the YOLO series algorithm is realized at the edge layer. In addition, the average power consumption is merely 16.8W. Further, we intend to adopt edge YOLO to a wider range of Edge computing devices and integrate the system into more ITS platforms and scenarios.

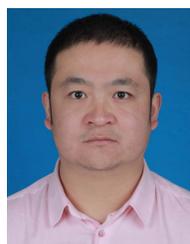

**Siyuan Liang** received the B.S. degree from the Xi'an University of Posts and Telecommunications, Xi'an, China, in 2006, and the Ph.D. degree from the Beijing University of Posts and Telecommunications, Beijing, China, in 2012. He is currently an Associate Professor with the School of Communications and Information Engineering, Xi'an University of Posts and Telecommunications. His main research interests include visual positioning, integrated navigation, and automated mission planning of unmanned systems.

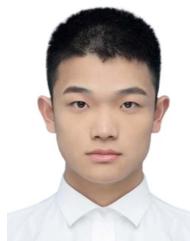

**Hao Wu** (Student Member, IEEE) is currently pursuing the B.S. degree from the Xi'an University of Posts and Telecommunications, Xi'an, China. He is a member of the Shaanxi Provincial Key Laboratory of Information Communication Network and Security. His main research interests include deep learning, image tracking, and unmanned system perception.

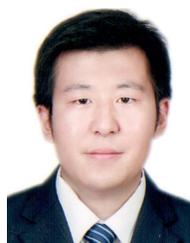

**Li Zhen** (Member, IEEE) received the M.S. degree in circuit and system from the Xi'an University of Posts and Telecommunications, Xi'an, China, in 2012, and the Ph.D. degree in communication and information systems from Xidian University, Xi'an, in 2018. He is currently a Lecturer with the School of Communication and Information Engineering, Xi'an University of Posts and Telecommunications. His research interests include 6G wireless communications, satellite mobile communications, weak signal detection and processing, and random multiple access.

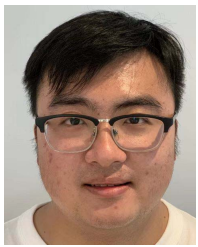

**Qiaozhi Hua** received the B.E. degree in electrical communication from the Wuhan University of Science and Technology, in 2011, and the M.S. and Ph.D. degrees from Waseda University in 2015 and 2019, respectively. He is currently a Lecturer with the Computer School, Hubei University of Arts and Science, Hubei, China. His research interests include game theory and wireless communications.

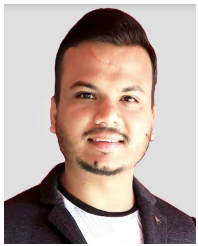

**Sahil Garg** (Member, IEEE) received the Ph.D. degree from the Thapar Institute of Engineering and Technology in 2018. He is currently a Research Associate at the Resilient Machine Learning Institute (ReMI) in correlation with École de Technologie Supérieure (ETS), Montréal. Prior to this, he worked as a Post-Doctoral Research Fellow at ETS and a MITACS Researcher at Ericsson, Montréal. He has many research contributions in the areas of machine learning, big data analytics, security and privacy, the Internet of Things, and cloud computing. He has over 80 publications in highly ranked journals and conferences, including more than 50 top-tier journal articles and more than 30 respected conference papers. He has been awarded the 2021 IEEE SYSTEMS JOURNAL Best Paper Award, the 2020 IEEE TCSC Award for Excellence in Scalable Computing (Early Career Researcher), and the IEEE ICC 2018 Best Paper Award. In addition, he also serves as the Workshops and Symposia Officer for the IEEE ComSoc Emerging Technology Initiative on Aerial Communications. He is currently a Managing Editor of Springer's *Human-Centric Computing and Information Sciences* journal. He is also an Associate Editor of *IEEE Network*, IEEE TRANSACTIONS ON INTELLIGENT TRANSPORTATION SYSTEMS, Elsevier's *Applied Soft Computing*, and Wiley's *International Journal on Communication Systems*.

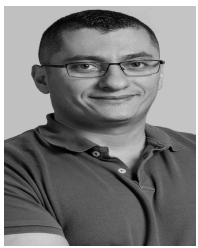

**Georges Kaddoum** (Member, IEEE) received the bachelor's degree in electrical engineering from the École Nationale Supérieure de Techniques Avancées (ENSTA Bretagne), Brest, France, the M.S. degree in telecommunications and signal processing (circuits, systems, and signal processing) from the Université de Bretagne Occidentale and Telecom Bretagne (ENSTB), Brest, in 2005, and the Ph.D. degree (Hons.) in signal processing and telecommunications from the National Institute of Applied Sciences (INSA), University of Toulouse, Toulouse, France, in 2009. Since 2010, he has been a Scientific Consultant in the field of space and wireless telecommunications for several U.S. and Canadian companies. He is currently a Full Professor and the Tier 2 Canada Research Chair with the École de Technologie Supérieure (ÉTS), Université du Québec, Montréal, Canada. He has authored or coauthored more than 150 journal and conference papers and has two pending patents. His recent research interests include mobile communication systems, modulations, security, and space communications and navigation.

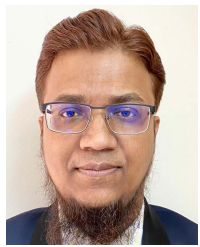

**Mohammad Mehedi Hassan** (Senior Member, IEEE) received the Ph.D. degree in computer engineering from Kyung Hee University, South Korea, in 2011. He is currently a Professor with the Information Systems Department, College of Computer and Information Sciences (CCIS), King Saud University (KSU), Riyadh, Saudi Arabia. He is one of the top Computer Scientist in Saudi Arabia as well. He has authored or coauthored around more than 260 publications. His research interests include cloud computing, body sensor networks, deep learning, mobile cloud, smart computing, wireless sensor networks, 5G networks, and social networks. He was a recipient of a number of awards. He is one of the top 2% scientists of the world in networking and telecommunication field.

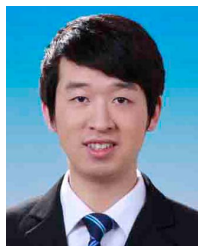

**Keping Yu** (Member, IEEE) received the M.E. and Ph.D. degrees from the Graduate School of Global Information and Telecommunication Studies, Waseda University, Tokyo, Japan, in 2012 and 2016, respectively. He was a Research Associate and a Junior Researcher with the Global Information and Telecommunication Institute, Waseda University, from 2015 to 2019 and from 2019 to 2020, where he is currently a Researcher. He has hosted and participated in more than ten projects, is involved in many standardization activities organized by ITU-T and ICNRG of IRTF, and has contributed to ITU-T Standards Y.3071 and Supplement 35. His research interests include smart grids, information-centric networking, the Internet of Things, blockchain, and information security. Moreover, he has served as a TPC Member of more than ten international conferences, including the ITU Kaleidoscope, IEEE VTC, IEEE CCNC, and IEEE WCNC. He was the Chair of the IEEE/CIC ICCC 2nd EBTSRA Workshop, the General Co-Chair and the Publicity Co-Chair of the IEEE VTC2020-Spring EBTSRA Workshop, the TPC Co-Chair of the SCML2020, the Local Chair of the MONAMI 2020, the Session Co-Chair of the CcS2020, and the Session Chair of the ITU Kaleidoscope 2016. He has been a Lead Guest Editor of *Sensors*, *Peer-to-Peer Networking and Applications*, and *Energies*, and a Guest Editor of *IEICE Transactions on Information and Systems*, *Intelligent Automation and Soft Computing*, and *Computer Communications*. He is an Editor of the IEEE OPEN JOURNAL OF VEHICULAR TECHNOLOGY.